\documentclass[10pt,twocolumn,letterpaper]{article}

\usepackage[pagenumbers]{wacv} 

\usepackage{graphicx}
\usepackage{amsmath}
\usepackage{amssymb}
\usepackage{booktabs}

\usepackage{multirow}
\usepackage{rotating}
\usepackage{makecell}
\usepackage{pifont}

\usepackage{xcolor,colortbl}
\usepackage[pagebackref,breaklinks,colorlinks]{hyperref}

\usepackage[capitalize]{cleveref}
\crefname{section}{Sec.}{Secs.}
\Crefname{section}{Section}{Sections}
\Crefname{table}{Table}{Tables}
\crefname{table}{Tab.}{Tabs.}

\def\confName{WACV}
\def\confYear{2025}

\usepackage{xspace}

\newcommand\minisection[1]{\vspace{2mm}\noindent \textbf{#1}}

\begin{document}

\title{TokenBinder: Text-Video Retrieval with One-to-Many Alignment Paradigm}

\author{Bingqing Zhang$^{1,2*}$, Zhuo Cao$^1$\thanks{Equal Contribution}, Heming Du$^1$, Xin Yu$^1$, Xue Li$^1$, Jiajun Liu$^{2,1\dag}$, Sen Wang$^1\thanks{Corresponding Authors}$\\
\\ $^1$ {The University of Queensland, Australia} \\
$^2$ {CSIRO Data61, Australia}\\
{\tt\small \{bingqing.zhang, william.cao, heming.du, xin.yu\}@uq.edu.au, xueli@eesc.uq.edu.au}\\
{\tt\small jiajun.liu@csiro.au, sen.wang@uq.edu.au}
}

\maketitle

\begin{abstract}
Text-Video Retrieval (TVR) methods typically match query-candidate pairs by aligning text and video features in coarse-grained, fine-grained, or combined (coarse-to-fine) manners. However, these frameworks predominantly employ a one(query)-to-one(candidate) alignment paradigm, which struggles to discern nuanced differences among candidates, leading to frequent mismatches. Inspired by Comparative Judgement in human cognitive science, where decisions are made by directly comparing items rather than evaluating them independently, we propose TokenBinder. This innovative two-stage TVR framework introduces a novel one-to-many coarse-to-fine alignment paradigm, imitating the human cognitive process of identifying specific items within a large collection. Our method employs a Focused-view Fusion Network with a sophisticated cross-attention mechanism, dynamically aligning and comparing features across multiple videos to capture finer nuances and contextual variations. Extensive experiments on six benchmark datasets confirm that TokenBinder substantially outperforms existing state-of-the-art methods. These results demonstrate its robustness and the effectiveness of its fine-grained alignment in bridging intra- and inter-modality information gaps in TVR tasks. Code is avaliable at https://github.com/bingqingzhang/TokenBinder.
\end{abstract}

\begin{figure}
    \centering
    \includegraphics[width=0.5\textwidth]{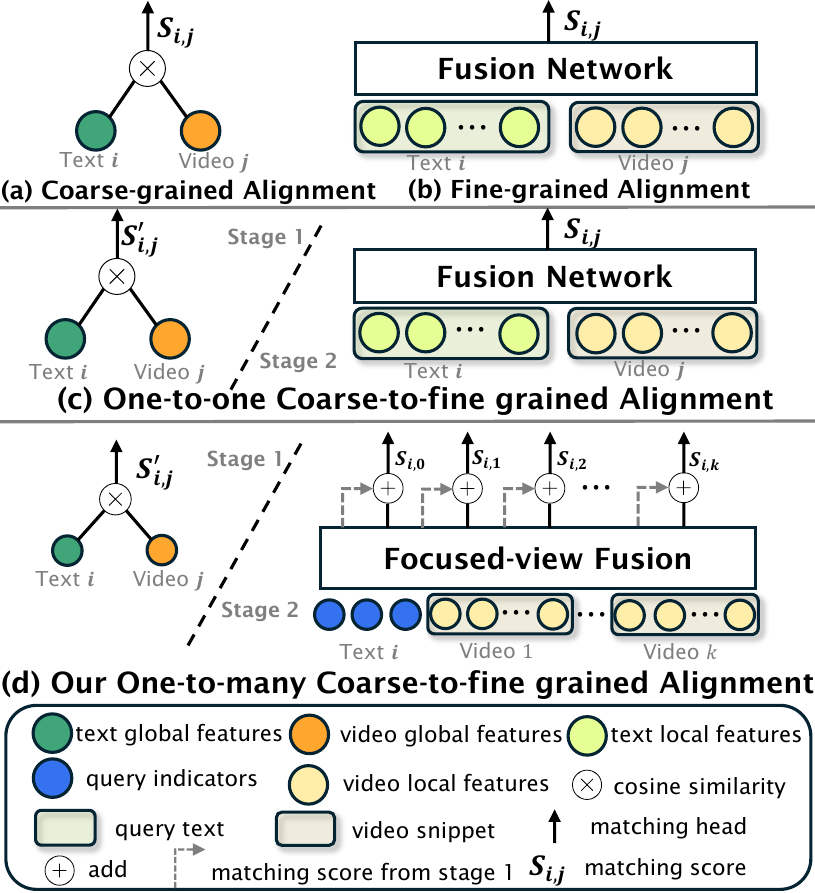}
    \caption{\textbf{An illustration of alignment methods in text-video retrieval}, categorized into three types: coarse-grained (a), fine-grained (b), and coarse-to-fine grained alignment (c). Traditional methods typically employ a one-to-one alignment paradigm. In contrast, we introduce a one-to-many coarse-to-fine grained alignment approach, allowing each query to be compared with multiple video candidates (d). This method facilitates mining differences among candidates to achieve enhanced retrieval effectiveness.}
    \label{fig:alignment_illustrate}
\end{figure}

\section{Introduction}
\label{sec:intro}

Recent developments have positioned the alignment of videos with natural language as a pivotal area of research. Text-Video Retrieval (TVR), a key task in direct cross-modal alignment, seeks to pair texts or videos with the most relevant counterparts within a multimedia database~\cite{camoe,clip_straight,clip2tv,clip4clip,clipvip,xpool}. This task significantly bridges the visual-textual information gap, enhancing both efficiency and accuracy in multimedia database searches.

A fundamental challenge in Text-Video Retrieval (TVR) is achieving effective text-video alignment. Initial studies~\cite{Torabi2016LearningLE,Kiros2014UnifyingVE,Kaufman2016TemporalTA,Zhang2018CrossModalAH} introduced various intensive alignment mechanisms. The emergence of the large-scale text-image pretrained model CLIP~\cite{clip} has substantially propelled advancements in TVR. Utilizing CLIP, Luo \etal~\cite{clip4clip} and subsequent studies~\cite{clipvip, clip2tv, prompt_switch} have implemented coarse-grained alignment methods. These methods aggregate both video and text features into a singular global token and process them using simple similarity functions (see Fig~\ref{fig:alignment_illustrate}a).
Further research~\cite{violet, xclip, xpool} has focused on developing hierarchical fusion networks that segment text and video features into multiple levels for fine-grained alignment; this includes breaking texts into words, phrases, and sentences, and videos into regions, frames, and snippets (see Fig~\ref{fig:alignment_illustrate}b). Additionally, recent studies~\cite{ucofia,eercf} have further integrated both coarse-grained and fine-grained alignment methods, proposing a two-stage coarse-to-fine grained framework to enhance the speed and quality of TVR (see Fig~\ref{fig:alignment_illustrate}c).

Despite these advancements, existing TVR methods predominantly utilize a one-to-one alignment paradigm, which often fails to distinguish nuanced differences among the most similar candidates that closely match the text query. Inspired by the Comparative Judgement in human cognitive science where decisions are made by directly comparing items rather than evaluating them independently \cite{pollitt2012method}, we propose a novel one-to-many coarse-to-fine alignment paradigm that mirrors human cognitive processes in identifying and distinguishing objects from a large dataset. This involves a three-step process: analyzing key components of the text query, quickly filtering videos in a coarse-grained manner, and then conducting detailed comparisons among candidates to identify the most relevant video. For instance, when given a query such as ``An old car with the logo PACOS is racing against another car," humans naturally focus on key details like ``old car", ``logo PACOS", and ``race", initially filtering for videos showcasing racing cars, and subsequently differentiating based on these specific aspects.

Building on this concept, we introduce TokenBinder, an innovative text-video retrieval method that integrates coarse-grained and fine-grained re-ranking within an end-to-end trainable framework. TokenBinder encapsulates the three-step re-ranking process: query-binding, broad-view retrieval (stage 1), and focused-view retrieval (stage 2). Initially, the query encoder is enhanced with additional tokens termed query-indicators which encapsulate key query information, enriching the significant details. During the broad-view retrieval, a query indicator serves as a global feature aggregator, calculating cosine similarity with all entries in the database to establish a preliminary ranking of candidates. This process is refined during the focused-view retrieval, where a Focused-view Fusion Network employs a cross-attention Transformer to align all candidate local features with the remaining query indicators. A Multilayer Perceptron (MLP) then calculates similarity logits, integrating these with scores from the initial retrieval to determine the final similarity scores (see Fig. \ref{fig:alignment_illustrate}d). By employing this one-to-many alignment paradigm, TokenBinder simultaneously compares all candidates, deriving relative similarity scores that reflect more nuanced distinctions. During training, the model is optimized using contrastive loss for broad-view and cross-entropy loss for focused-view retrieval, ensuring robust learning of both coarse and fine details.

The cornerstone of our framework, TokenBinder, is the strategic use of query-indicators. These indicators initially bind critical query-modality information during broad-view retrieval and subsequently link to candidate features during focused-view retrieval. Functioning as dynamic pools of abstract information, they effectively encapsulate essential features from both texts and videos. In fact, these indicators facilitate a reduction of cognitive load \cite{sweller1988cognitive} in Comparative Judgement theory, where larger units of information can be compressed into smaller, more manageable units.

To demonstrate the efficacy of TokenBinder, we conducted extensive testing across several standard text-video retrieval datasets. Employing TokenBinder with a ViT-B/32 model, we achieved remarkable R@1 scores: 52.5\% on MSRVTT-1k~\cite{xu2016msr-vtt}, 48.3\% on MSVD~\cite{msvd}, 48.3\% on ActivityNet~\cite{caba2015activitynet}, 62.7\% on VATEX~\cite{Wang2019VaTeXAL}, and 48.2\% on DiDeMO~\cite{hendricks18emnlp}. These results not only affirm the robustness of TokenBinder but also indicate its superiority over current state-of-the-art CLIP-based methods by margins of 2.3\%, 1.1\%, 2.1\%, 1.3\%, and 0.8\% respectively.

\section{Related Work}
\label{sec:rela}
\minisection{Text-Video Retrieval.}
Text-video retrieval ~\cite{clip_straight, clip4clip, Yu2018AJS, teachtext, Yang2021TACoTC, Chen2020FineGrainedVR, Lu2023UniAdapterUP, Shu2022MaskedCP}, a key task in vision-language processing, involves two main subtasks: text-to-video (\textit{t2v}) and video-to-text (\textit{v2t}) retrieval. Historically, text-to-video retrieval  has been the primary focus, with numerous models emphasizing this as their main result. The introduction of the Attention mechanism \cite{vaswani2017attention} has significantly influenced this field, leading to advancements such as MMT \cite{MMT}, which integrates temporal and cross-modal cues, and SupportSet \cite{support_set}, which uses generative methods to cluster related samples.

A major milestone in text-video retrieval was the introduction of CLIP \cite{rao2022denseclip}, which significantly simplified and enhanced the retrieval process. Luo et al. \cite{clip4clip} innovated by processing videos as sequences of frames through CLIP’s vision encoders, employing mean pooling to generate features. This streamlined approach not only simplified the processing but also led to notable performance improvements, catalyzing extensive subsequent research. In encoder extraction \cite{ts2net, prompt_switch, clipvip, centerclip, Wang2021ObjectawareVP}, efforts have focused on developing more precise and comprehensive video and text encoders. In modality alignment \cite{eercf, ucofia, xclip, xpool, Chen2023TaggingBA}, researchers have explored various strategies to better align modalities. Training optimization \cite{Yang2021TACoTC, clipvip, Zhang2023MultieventVR} has targeted enhancements in training methodologies and loss functions, while normalization \cite{camoe, Park2022NormalizedCL, Bogolin2021CrossMR} has sought to improve the effectiveness of the similarity matrix. Our framework, TokenBinder, builds on these advancements, focusing specifically on modality alignment to foster more effective interactions between text and video modalities, thus contributing to the ongoing evolution of the field.

\begin{figure*}[htbp]
    \centering
    \includegraphics[width=1.0\linewidth]{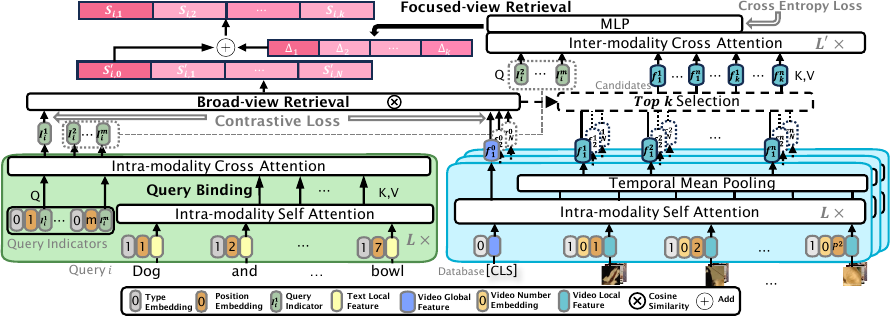}
    \caption{\textbf{TokenBinder Framework for Text-Video Retrieval.} The diagram showcases the complete workflow of our dual-stage retrieval system. Initially, the query is processed using intra-modality cross attention to bind significant query indicators with textual features, shown in the green section. The broad-view retrieval then ranks video candidates based on their global features using cosine similarity and contrastive loss, illustrated in middle section. The top-ranked candidates are further refined in the focused-view retrieval through inter-modality cross attention and MLP-based similarity scoring, as depicted in the top section. This process ensures comprehensive text-video alignment and optimizes retrieval accuracy.}
    \label{fig:framework}
\end{figure*}

\minisection{Multimodal Alignment in Text-Video Retrieval.}
Multimodal alignment is critical in text-video retrieval, with methods ranging from coarse-grained to fine-grained, and a combination approach of coarse-to-fine grained alignment. Coarse-grained alignment combines text and video into a global token and uses simple metrics like cosine similarity for evaluation. Fine-grained alignment aligns at the level of segments, frames, and patches for video and sentences, phrases, and words for text. Despite their efficacy, these approaches often miss either detailed or high-level information. The coarse-to-fine methods \cite{ucofia, eercf} aim to balance these scales by integrating features across different granularity. However, most remain limited to one-to-one alignments. In contrast, our TokenBinder framework adopts a one-to-many alignment paradigm, considering candidate videos simultaneously and providing a nuanced understanding of cross-modal relationships.

\section{Method}

This section begins with an overview of the foundational concepts and notations used in text-video retrieval. Subsequently, we illustrate an overall framework of TokenBinder, outlining the primary inputs and procedural steps involved. Then we describe the textual and visual expressions in TokenBinder. The methodology is divided into two distinct stages of retrieval within TokenBinder: the Broad-view Retrieval and the Focused-view Retrieval. Each stage is discussed in detail. Finally, we present the dual-stage training strategy employed to optimize retrieval performance.

\subsection{Preliminaries}
In the field of Text-Video Retrieval (TVR), we primarily engage with two subtasks: text-to-video retrieval (\textit{t2v}) and video-to-text retrieval (\textit{v2t}). The goal of \textit{t2v} is to identify the video within a database that most accurately aligns with a specified text query. Conversely, \textit{v2t} retrieval focuses on finding the textual description that best matches a given video. Although these tasks are symmetrical in nature, \textit{t2v} has historically garnered more research focus. Therefore, our discussion and subsequent results predominantly address the \textit{t2v} task.

In the CLIP-based framework, a text sequence $\mathcal{T}$ is composed of $M$ tokens, expressed as $\mathcal{T} = \{t_1, t_2, ..., t_M\ | t_i \in \mathbb{R}^C \}$. Similarly, a video $\mathcal{V}$ is represented by a sequence of frames $\mathcal{V} = \{f_{[cls]},f_1, f_2, ..., f_T\}$, $f_{[cls]}$ denotes the global feature of the video $\mathcal{V}$, $T$ is the number of frame, each frame $f_i$ being a tensor in $\mathbb{R}^{H \times W \times C}$ with $H$, $W$, and $C$ representing height, width, and channel count, respectively. To facilitate feature extraction, the CLIP framework dissects each frame $f_i$ into $P^2$ patches of size $P \times P$, thus described by $\{f_{i}^{1}, f_{i}^{2}, ..., f_{i}^{P^{2}} | f_{i}^{j} \in \mathbb{R}^C \}$.

\subsection{Overall Framework}
\label{sec:overallframework}
As depicted in Figure~\ref{fig:framework}, our proposed framework, TokenBinder, introduces a novel two-stage CLIP-based system designed to optimize text-video retrieval. The TokenBinder incorporates a three-step re-ranking process: query-binding, broad-view retrieval (stage 1), and focused-view retrieval (stage 2). The process begins with an advanced query encoding mechanism, wherein special tokens, termed query indicators, are embedded within the query encoder. These indicators are crucial for capturing key information, thereby dynamically enhancing the query's ability to interact more effectively with video content across different stages. This enhancement not only facilitates a nuanced interaction with video features but also lays the groundwork for a sophisticated retrieval process.

In the broad-view retrieval stage, TokenBinder leverages the enriched queries to conduct a global analysis of the video database. The first query indicator serves as an anchor, gathering global features from videos and establishing preliminary rankings based on cosine similarities. This initial phase effectively filters out less relevant video candidates, streamlining the selection process. The subsequent focused-view retrieval stage involves a deeper examination of a select group of top-ranked videos. Here, a Focused-view Fusion Network, equipped with a cross-attention Transformer mechanism, precisely aligns the remaining query indicators with the local features of these video candidates. The integration of detailed alignment with initial global insights is accomplished through a MLP, which calculates the final similarity scores, ultimately determining the most relevant video outputs.

This layered approach ensures that TokenBinder captures not only the broad thematic elements of the query but also explores the finer textual and visual nuances of the video content. The framework's capacity to dynamically adjust its focus during the retrieval process marks a significant advancement in text-video retrieval technology, resembling a conditional attention mechanism that optimizes for relevance and precision. Through these innovative features, TokenBinder sets a new benchmark in the field, offering a comprehensive method for engaging with complex multimedia content.

\subsection{Feature Representation in Text and Video}
Effectively representing text and video features is foundational in our TokenBinder framework. Utilizing the CLIP architecture as our base, we handle text and video feature representation distinctly. In line with leading studies such as \cite{clip4clip,centerclip, clip2tv}, video features are extracted using a Vision Transformer (ViT) \cite{vit}. Videos are decomposed into a series of frames, each enriched with three types of embedding (i.e. type embedding, video number embedding, and position embedding) to distinguish different tokens. Additionally, a [CLS] token is integrated to capture global video information comprehensively. Following the final ViT layer, a temporal mean pooling strategy \cite{eercf,xclip} is employed to significantly reduce the volume of local tokens while preserving critical video information. Formally, for a series of $i$-th frame patch features $\{ f^1_i, f^2_i, \ldots, f^{P^2}_i \}$, after temporal mean pooling we have $\{ f^1_i, f^2_i, \ldots, f^n_i \}$, where $n << P^2$ and the number is re-indexed.

Text features are enhanced by query indicators, which are initialized randomly and processed with text tokens. At each text encoder layer, tokens undergo a self-attention module from the CLIP encoder. The query binding process then aggregates information from these tokens into the query indicators. For query indicators at layer $l$: ${I_{1}^{l}, I_{2}^{l}, \ldots, I_{m}^{l} }$, the update mechanism is defined as:

\begin{equation}
    \begin{split}
        \{ I_{1}^{l}, I_{2}^{l}, \ldots, I_{m}^{l} \}  & = \theta ( Q := \{ I_{1}^{l-1}, I_{2}^{l-1}, \ldots, I_{m}^{l-1} \}; \\ & K, V := \{ t_{1}^{l-1}, t_{2}^{l-1}, \ldots, t_{M}^{l-1} \} ).
    \end{split}
\end{equation}
Here, $\theta$ represents the cross-attention layer within the text encoder, inserted after each self-attention module to dynamically capture relevant information, improving the query’s interaction with video content.

\subsection{Broad-view Retrieval}

Following the extraction of textual and visual representations, the broad-view retrieval phase employs global features from the text query and videos to compute cosine similarity scores rapidly. The query indicators, which bind all key information during the text encoding process, serve as the global features for the text query. Specifically, we utilize the first query indicator after binding $i$-th query information $I_{1,i}$ to represent global text features. These representations facilitate the calculation of similarity scores across the dataset. The broad-view similarity between a text query \( t_i \) and all videos in the dataset $N$ is calculated as follows:

\begin{equation}
    \begin{split}
        \mathcal{S^{'}}_{i} & = (s^{'}_{i,1}, s^{'}_{i,2}, \ldots, s^{'}_{i,N}) \\ & = \left( I_{1, i}^{\top} \cdot f_{1,[cls]}, I_{1, i}^{\top} \cdot f_{2,[cls]}, \ldots, I_{1, i}^{\top} \cdot f_{N,[cls]} \right).
    \end{split}
\end{equation}

Each similarity score \(s^{'}_{i,j}\) within the vector $\mathcal{S}^{'}_{i}$ represents the dot product between the global text representation $I_{1, i}^{\top}$ of query $i$ and the global video representation \(f_{j,[cls]}\) of each video. This approach quantifies the degree of relevance or similarity between the textual query and each video in the dataset, a crucial metric for effective retrieval.

\subsection{Focused-view Retrieval}

Following the initial phase of Broad-view Retrieval, we obtain preliminary retrieval scores. These initial scores, while effective for a broad matching, often do not capture the intricate details within videos, necessitating further refinement. This need leads to the Focused-view Retrieval.

In this stage, the scores \( \mathcal{S^{'}}_{i} \) are used to select the top-$k$ videos as focused candidates for more precise optimization. This selection is facilitated by the focused-view fusion module, which comprises a cross-attention Transformer and a Multilayer Perceptron (MLP) network. The cross-attention Transformer specifically targets the binding of local features from the top-$k$ video candidates into the remaining query indicators post the Broad-view Retrieval. These enriched query indicators now embody comprehensive information necessary for precise alignment between the query and video candidates. Consequently, the MLP network projects these indicators into an alignment space, treating the projection as a classification problem to directly output the classification scores for these k videos with a Softmax layer. The formal notation of the cross-attention Transformer process $\eta(\cdot)$ is expressed as:

\begin{equation}
    \begin{split}
        \{ I^{l}_{2}, I^{l}_{3}, \ldots, I^{l}_{m} \}  & = \eta ( Q := \{ I^{l-1}_{2}, I^{l-1}_{3}, \ldots, I^{l-1}_{m} \}; \\ & K, V = \{ f_{1}^1, f_{1}^2, \ldots, f_{k}^{1}, \ldots, f_{k}^n \} ).
    \end{split}
\end{equation}

To facilitate differentiable sampling from a discrete distribution, we employ the Gumbel-Softmax~\cite{Jang2016CategoricalRW,Maddison2016TheCD} to replace the traditional Softmax in the $\eta(\cdot)$ operation. Following this, an MLP is used to project the integrated information:

\begin{equation}
\{ \Delta_1, \Delta_i, \ldots, \Delta_{k} \} = \text{MLP} ( I_{2}, I_{3}, \ldots, I_{m} ).
\end{equation}

Ultimately, the refined similarity scores for a query $i$ in the stage-2 retrieval are calculated as:

\begin{equation}
\mathcal{S}_i = \left( s^{'}_{i,1} + \Delta_1, s^{'}_{i,2} + \Delta_2, \ldots, s^{'}_{i,k} + \Delta_{k} \right).
\end{equation}

\subsection{Broad- and Focused-view Supervision Loss}
\vspace{-0.5em}
During the training phase, we tailor specific optimization targets for the two stages of retrieval to supervise different query indicators, enhancing distinct aspects' performance.

In the first stage, we utilize a Contrastive Loss to align the first query indicator \(I_1\) in text with the global video token \(f^{[CLS]}\). Formally, for a batch of size \( B \), the contrastive loss functions for the \textit{t2v} and \textit{v2t} alignments are defined as follows:

\begin{equation}
\ell_{t \to v} = - \frac{1}{B} \sum_{i=1}^{B} \frac{e^{I_{1,i}^{\top} f^{[CLS]}_{i} / \tau}}{\sum_{j=1}^{B} e^{I_{1,i}^{\top} f_{j}^{[CLS]} / \tau}},
\end{equation}

\begin{equation}
\ell_{v \to t} = - \frac{1}{B} \sum_{i=1}^{B} \frac{e^{{f_{i}^{[CLS]}}^{\top} I_{1,i} / \tau}}{\sum_{j=1}^{B} e^{{f^{[CLS] }_{i}}^{\top} I_{1,j} / \tau}}.
\end{equation}
Here, \( \tau \) is the temperature parameter that scales the logits before applying the softmax function, effectively minimizing the distances between aligned pairs relative to other pairs in the batch.

For the focused view, the aim is to refine discrimination among closely related video candidates. This is achieved through a cross-entropy loss, which treats the scores projected by an MLP with a Softmax layer as class probabilities for a classification task. Given the distinct roles of video and sentence indicators, we apply supervision separately to each token type, denoted as \( \ell_{\text{focus},v} \) and \( \ell_{\text{focus},t} \).

The combined loss function, integrating both broad and focused supervision, is calculated as follows:

\begin{equation}
\ell = \left( \ell_{t \to v} + \ell_{v \to t} \right) / 2 + \left( \ell_{\text{focus}, v} + \ell_{\text{focus}, t} \right) / 2 .
\end{equation}

This formulation ensures that both the broad and focused objectives contribute equally to the overall training process, promoting robust learning across both modalities and enhancing detail within the representations.

\begin{table*}[htbp]
\centering
\footnotesize
\caption{Comparison of retrieval results on MSRVTT-1K-Test. The upper section details the performance of methods not employing CLIP, while the middle section delineates the outcomes for CLIP-based models utilizing a ViT-B/32 backbone. The lower section corresponds to results achieved with a ViT-B/16 backbone.}
\label{tab:msrvtt1k}
\setlength{\tabcolsep}{9.5pt}
\begin{tabular}{lcccccccccc}
\toprule[1pt]
\multicolumn{1}{l|}{\multirow{2}{*}{\textbf{Model}}} & \multicolumn{5}{c|}{\textbf{ \textit{t2v} Retrieval}}                                                                           & \multicolumn{5}{c}{\textbf{ \textit{v2t} Retrieval}}                                                                       \\
\multicolumn{1}{l|}{}                                & R@1  $\uparrow$                & R@5  $\uparrow$                 & R@10 $\uparrow$                & MdR      $\downarrow$            & \multicolumn{1}{c|}{MnR $\downarrow$}  & R@1    $\uparrow$       & R@5  $\uparrow$     & R@10  $\uparrow$   & MdR        $\downarrow$          & MnR  $\downarrow$                \\ \hline
\multicolumn{11}{c}{\textit{Non-CLIP}}                                                                                                                                                                                                                                                          \\ \hline
\multicolumn{1}{l|}{CE \cite{CE}}                              & 20.9                 & 48.8                 & 62.4                 & 6.0                  & \multicolumn{1}{c|}{28.2} & 20.6                 & 50.3                 & 64.0                 & 5.3                  & 25.1                 \\
\multicolumn{1}{l|}{MMT \cite{MMT}}                             & 26.6                 & 57.1                 & 69.6                 & 4.0                  & \multicolumn{1}{c|}{24.0} & 27.0                 & 57.5                 & 69.7                 & 3.7                  & 21.3                 \\
\multicolumn{1}{l|}{Support Set \cite{support_set}}                     & 27.4                 & 56.3                 & 67.7                 & 3.0                  & \multicolumn{1}{c|}{-}    & 26.6                 & 55.1                 & 67.5                 & 3.0                  & -                    \\
\multicolumn{1}{l|}{Frozen \cite{frozen}}                          & 31.0                 & 59.5                 & 70.5                 & 3.0                  & \multicolumn{1}{c|}{-}    & -                    & -                    & -                    & -                    & -                    \\
\multicolumn{1}{l|}{HiT \cite{hit}}                             & 30.7                 & 60.9                 & 73.2                 & 2.6                  & \multicolumn{1}{c|}{-}    & 32.1                 & 62.7                 & 74.1                 & 3.0                  & -                    \\
\multicolumn{1}{l|}{TeachText \cite{teachtext}}                           & 29.6                 & 61.6                 & 74.2                 & 3.0                  & \multicolumn{1}{c|}{-}    & 32.1                 & 62.7                 & 75.0                 & 3.0                  & -                    \\ \hline
\multicolumn{11}{c}{\textit{Backbone model: ViT-B/32}}                                                                                                                                                                                                                                          \\ \hline
\multicolumn{1}{l|}{CLIP-straight \cite{clip_straight}}                   & 31.2                 & 53.7                 & 64.2                 & 4.0                  & \multicolumn{1}{c|}{-}    & 27.2                 & 51.7                 & 62.6                 & 5.0                  & -                    \\
\multicolumn{1}{l|}{CLIP4Clip \cite{clip4clip}}                       & 44.5                 & 71.4                 & 81.6                 & 2.0                  & \multicolumn{1}{c|}{15.3} & 42.7                 & 70.9                 & 80.6                 & 2.0                  & 11.6                 \\
\multicolumn{1}{l|}{CAMoE \cite{camoe}}                           & 44.6                 & 72.6                 & 81.8                 & 2.0                  & \multicolumn{1}{c|}{13.3} & 45.1                 & 72.4                 & 83.1                 & 2.0                  & 10.0                 \\
\multicolumn{1}{l|}{CenterCLIP \cite{centerclip}}                      & 44.2                 & 71.6                 & 82.1                 & -                    & \multicolumn{1}{c|}{15.1} & 42.8                 & 71.7                 & 82.2                 & -                    & 11.1                 \\
\multicolumn{1}{l|}{CLIP2TV \cite{clip2tv}}                         & 45.6                 & 71.1                 & 80.8                 & 2.0                  & \multicolumn{1}{c|}{14.6} & 43.9                 & 70.9                 & 82.2                 & 2.0                  & 12.0                 \\
\multicolumn{1}{l|}{X-Pool \cite{xpool}}                          & 46.9                 & 72.8                 & 82.2                 & 2.0                  & \multicolumn{1}{c|}{14.3} & 44.4                 & 73.3                 & 84.0                 & -                    & 9.0                  \\
\multicolumn{1}{l|}{X-CLIP \cite{xclip}}                          & 46.1                 & 73.0                 & 83.1                 & 2.0                  & \multicolumn{1}{c|}{13.2} & 46.8                 & 73.3                 & 84.0                 & 2.0                  & 9.1                  \\
\multicolumn{1}{l|}{TS2-Net \cite{ts2net}}                         & 47.0                 & 74.5                 & 83.8                 & 2.0                  & \multicolumn{1}{c|}{13.0} & 45.3                 & 74.1                 & 83.7                 & 2.0                  & 9.2                  \\
\multicolumn{1}{l|}{DRL \cite{drl}}                             & 47.4                 & 74.6                 & 83.8                 & 2.0                  & \multicolumn{1}{c|}{-}    & 45.3                 & 73.9                 & 83.3                 & 2.0                  & -                    \\
\multicolumn{1}{l|}{UCOFIA \cite{ucofia}}                          & 49.4                 & 72.1                 & 83.5                 & 2.0                  & \multicolumn{1}{c|}{12.9} & 47.1                 & 74.3                 & 83.0                 & 2.0                  & 11.4                 \\
\multicolumn{1}{l|}{PromptSwitch \cite{prompt_switch}}                    & 46.1                 & 72.8                 & 81.8                 & -                    & \multicolumn{1}{c|}{14.4} & 44.8                 & 73.7                 & 82.4                 & -                    & 9.9                  \\
\multicolumn{1}{l|}{EERCF \cite{eercf}}                           & 47.8                 & 74.1                 & 84.1                 & -                    & \multicolumn{1}{c|}{-}    & 44.7                 & 74.2                 & 83.9                 & -                    & -                    \\
\multicolumn{1}{l|}{CLIP-ViP \cite{clipvip}}                         & 50.2                 & 74.8                 & \textbf{84.2}                 & 1.0                  & \multicolumn{1}{c|}{13.9} & 48.1                 & 75.2                 & \textbf{84.6}                 & 2.0                  & 9.5                  \\
\multicolumn{1}{l|}{DGL \cite{dgl}}                             & 45.8                 & 69.3                 & 79.4                 & -                    & \multicolumn{1}{c|}{16.3} & 43.5                 & 70.5                 & 80.7                 & -                    & 13.1                 \\
\multicolumn{1}{l|}{ProST \cite{prost}}                           & 48.2                 & 74.6                 & 83.4                 & 2.0                  & \multicolumn{1}{c|}{\textbf{12.4}} & 46.3                 & 74.2                 & 83.2                 & 2.0                  & \textbf{8.7}                  \\ \hline
\rowcolor[HTML]{EFEFEF}\multicolumn{1}{l|}{\textbf{Ours} }                             & \textbf{52.5} & \textbf{75.3} & \textbf{84.2} & \textbf{1.0} & 12.9     & \textbf{52.0} & \textbf{75.5} & \textbf{84.6} & \textbf{1.0}{} & 9.1 \\ \hline
\multicolumn{11}{c}{\textit{Backbone model: ViT-B/16}}                                                                                                                                                                                                                                          \\ \hline
\multicolumn{1}{l|}{CenterCLIP \cite{centerclip}}                      & 48.4                 & 73.8                 & 82.0                 & 2.0                  & \multicolumn{1}{c|}{13.8} & 47.7                 & 75.0                 & 83.3                 & 2.0                  & 10.2                 \\
\multicolumn{1}{l|}{CLIP2TV \cite{clip2tv}}                         & 49.3                 & 74.7                 & 83.6                 & 2.0                  & \multicolumn{1}{c|}{13.5} & 46.9                 & 75.0                 & 85.1                 & 2.0                  & 10.0                 \\
\multicolumn{1}{l|}{DRL \cite{drl} }                             & 50.2                 & 76.5                 & \textbf{84.7}                 & 2.0                  & \multicolumn{1}{c|}{-}    & 48.9                 & 76.3                 & 85.4                 & 2.0                  & -                    \\
\multicolumn{1}{l|}{EERCF \cite{eercf}}                           & 49.9                 & 76.5                 & 84.2                 & -                    & \multicolumn{1}{c|}{-}    & 47.8                 & 75.3                 & 84.2                 & -                    & -                    \\
\multicolumn{1}{l|}{CLIPVIP \cite{clipvip}}                         &     53.2                 &         77.4             &     84.6                 &        \textbf{1.0}              & \multicolumn{1}{c|}{11.8}     &           50.6           &             78.8         &  85.6                    &      \textbf{1.0}                &         7.8             \\
\multicolumn{1}{l|}{DGL \cite{dgl} }                             & 48.3                 & 71.8                 & 80.6                 & -                    & \multicolumn{1}{c|}{13.4} & 45.7                 & 74.0                 & 82.9                 & -                    & 10.9                 \\
\multicolumn{1}{l|}{ProST \cite{prost}}                           & 49.5                 & 75.0                 & 84.0                 & 2.0                  & \multicolumn{1}{c|}{11.7} & 48.0                 & 75.9                 & 85.2                 & 2.0                  & 8.3                  \\ \hline
\rowcolor[HTML]{EFEFEF}\multicolumn{1}{l|}{\textbf{Ours} }                           & \textbf{54.6} & \textbf{77.5} & \textbf{ 84.7} & \textbf{1.0} & \textbf{11.7}     & \textbf{52.4} & \textbf{79.0} & \textbf{85.8} & \textbf{1.0} & \textbf{7.7} \\ \bottomrule[1pt]
\end{tabular}
\end{table*}

\begin{table*}[htbp]
\centering
\small
\caption{Comparison results on ActivityNet, DiDeMo, VATEX, MSVD and LSMDC dataset. CLIP-based Models use ViT-B/32 backbone.}
\label{tab:multi_dataset}
\setlength{\tabcolsep}{2pt}
\begin{tabular}{lccccccccccccccc}
\toprule[1pt]
\multicolumn{1}{l|}{\multirow{2}{*}{\textbf{Method}}} & \multicolumn{3}{c|}{\textbf{ActivityNet}}                         & \multicolumn{3}{c|}{\textbf{DiDeMo}}                               & \multicolumn{3}{c|}{\textbf{VATEX}}                               & \multicolumn{3}{c|}{\textbf{MSVD}}                                & \multicolumn{3}{c}{\textbf{LSMDC}}            \\
\multicolumn{1}{l|}{}                                 & R@1   $\uparrow$        & R@5   $\uparrow$        & \multicolumn{1}{c|}{MnR $\downarrow$}          & R@1   $\uparrow$        & R@5   $\uparrow$     & \multicolumn{1}{c|}{MnR $\downarrow$}           & R@1   $\uparrow$     & R@5  $\uparrow$   & \multicolumn{1}{c|}{MnR $\downarrow$}          & R@1   $\uparrow$        & R@5   $\uparrow$      & \multicolumn{1}{c|}{MnR $\downarrow$}          & R@1 $\uparrow$   & R@5  $\uparrow$   & MnR   $\downarrow$ \\ \hline
\multicolumn{16}{c}{Non-CLIP}                                                                                                                                                                                                                                                                                                                                                          \\ \hline
\multicolumn{1}{l|}{CE \cite{CE}}                               & 18.2          & 47.7          & \multicolumn{1}{c|}{23.1}         & 16.1          & 41.1          & \multicolumn{1}{c|}{43.7}          & -             & -             & \multicolumn{1}{c|}{-}            & 19.8          & 49.0          & \multicolumn{1}{c|}{23.1}         & -             & -             & -             \\
\multicolumn{1}{l|}{Support Set \cite{support_set}}                      & 29.2          & 61.6          & \multicolumn{1}{c|}{-}            & -             & -             & \multicolumn{1}{c|}{-}             & 45.9          & 82.4          & \multicolumn{1}{c|}{-}            & 28.4          & 60.0          & \multicolumn{1}{c|}{-}            & -             & -             & -             \\
\multicolumn{1}{l|}{Frozen \cite{frozen}}                           & -             & -             & \multicolumn{1}{c|}{-}            & 34.6          & 65.0          & \multicolumn{1}{c|}{-}             & -             & -             & \multicolumn{1}{c|}{-}            & 33.7          & 64.7          & \multicolumn{1}{c|}{-}            & 15.0          & 30.8          & -             \\
\multicolumn{1}{l|}{HiT \cite{hit}}                              & 29.6          & 60.7          & \multicolumn{1}{c|}{-}            & -             & -             & \multicolumn{1}{c|}{-}             & -             & -             & \multicolumn{1}{c|}{-}            & -             & -             & \multicolumn{1}{c|}{-}            & 14.0          & 31.2          & -             \\
\multicolumn{1}{l|}{TeachText\cite{teachtext}}                        & 25.0          & 58.7          & \multicolumn{1}{c|}{-}            & 21.6          & 48.6          & \multicolumn{1}{c|}{-}             & -             & -             & \multicolumn{1}{c|}{-}            & 22.1          & 52.2          & \multicolumn{1}{c|}{-}            & 17.2          & 35.6          & -             \\ \hline
\multicolumn{16}{c}{CLIP-based}                                                                                                                                                                                                                                                                                                                                                        \\ \hline
\multicolumn{1}{l|}{CLIP4Clip \cite{clip4clip}}                        & 40.5          & 72.4          & \multicolumn{1}{c|}{7.4}          & 43.4          & 70.2          & \multicolumn{1}{c|}{17.5}          & 55.9          & 89.2          & \multicolumn{1}{c|}{3.9}          & 46.2          & 76.1          & \multicolumn{1}{c|}{10.0}         & 22.6          & 41.0          & 61.0          \\
\multicolumn{1}{l|}{CenterCLIP \cite{centerclip}}                       & 43.9          & 75.3          & \multicolumn{1}{c|}{7.0}          & -             & -             & \multicolumn{1}{c|}{-}             & -             & -             & \multicolumn{1}{c|}{-}            & 47.3          & 76.8          & \multicolumn{1}{c|}{9.7}          & 21.4          & 39.7          & 55.9          \\
\multicolumn{1}{l|}{CLIP2TV \cite{clip2tv}}                          & 40.8          & 72.9          & \multicolumn{1}{c|}{6.5}          & 43.9          & 70.5          & \multicolumn{1}{c|}{16.6}          & 61.4          & 90.6          & \multicolumn{1}{c|}{3.7}          & 46.3          & 76.1          & \multicolumn{1}{c|}{10.0}         & -             & -             & -             \\
\multicolumn{1}{l|}{X-CLIP \cite{xclip}}                           & 44.3          & 74.1          & \multicolumn{1}{c|}{-}            & 45.2          & 74.0          & \multicolumn{1}{c|}{14.6}          & -             & -             & \multicolumn{1}{c|}{-}            & 47.1          & 77.8          & \multicolumn{1}{c|}{9.5}          & 23.3          & 43.0          & 56.0          \\
\multicolumn{1}{l|}{TS2-Net \cite{ts2net}}                          & 41.6          & 73.6          & \multicolumn{1}{c|}{8.4}          & 41.8          & 71.6          & \multicolumn{1}{c|}{14.8}          & 59.1          & 90.0          & \multicolumn{1}{c|}{3.5}          & -             & -             & \multicolumn{1}{c|}{-}            & -             & -             & -             \\
\multicolumn{1}{l|}{UCOFIA \cite{ucofia}}                           & 45.7          & \textbf{76.0} & \multicolumn{1}{c|}{6.6}          & 46.5          & 74.8          & \multicolumn{1}{c|}{\textbf{13.1}} & 61.1          & 90.5          & \multicolumn{1}{c|}{3.4}          & 47.4          & 77.6          & \multicolumn{1}{c|}{9.6}          & -             & -             & -             \\
\multicolumn{1}{l|}{EERCF \cite{eercf}}                            & 43.1          & 74.5          & \multicolumn{1}{c|}{-}            & -             & -             & \multicolumn{1}{c|}{-}             & 62.6          & \textbf{91.5} & \multicolumn{1}{c|}{-}            & 47.0          & 77.5          & \multicolumn{1}{c|}{-}            & -             & -             & -             \\
\multicolumn{1}{l|}{DGL \cite{dgl}}                              & 43.1          & 72.3          & \multicolumn{1}{c|}{8.6}          & -             & -             & \multicolumn{1}{c|}{-}             & 57.3          & 87.1          & \multicolumn{1}{c|}{4.1}          & -             & -             & \multicolumn{1}{c|}{-}            & -             & -             & -             \\
\multicolumn{1}{l|}{ProST \cite{prost}}                            & -             & -             & \multicolumn{1}{c|}{-}            & 44.9          & 72.7          & \multicolumn{1}{c|}{13.7}          & 60.6          & 90.5          & \multicolumn{1}{c|}{3.4}          & -             & -             & \multicolumn{1}{c|}{-}            & -             & -             & -             \\
\multicolumn{1}{l|}{CLIP-ViP \cite{clipvip}}                         & 45.9          & 74.2          & \multicolumn{1}{c|}{8.1}          & 47.4          & 75.2          & \multicolumn{1}{c|}{13.5}          & 60.9          & 89.5          & \multicolumn{1}{c|}{3.4}          & 47.2          & 77.1          & \multicolumn{1}{c|}{9.5}          & 24.6          & 44.5          & 53.8          \\ \hline
\rowcolor[HTML]{EFEFEF} \multicolumn{1}{l|}{Ours}                             & \textbf{46.3} & 74.2          & \multicolumn{1}{c|}{\textbf{7.3}} & \textbf{48.2} & \textbf{76.7} & \multicolumn{1}{c|}{\textbf{13.1}} & \textbf{62.7} & 90.5          & \multicolumn{1}{c|}{\textbf{3.1}} & \textbf{48.3} & \textbf{78.3} & \multicolumn{1}{c|}{\textbf{8.9}} & \textbf{25.0} & \textbf{44.7} & \textbf{51.1} \\ \bottomrule[1pt]
\end{tabular}

\end{table*}

\section{Experiments}
\subsection{Datasets}
We evaluate our proposed model, TokenBinder, on 6 widely recognized video text retrieval datasets: MSRVTT, DiDeMo, ActivityNet, VATEX, MSVD and LSMDC. 

\textbf{MSR-VTT~\cite{xu2016msr-vtt}} is one of the most extensively utilized in the field. The more widely adopted split in recent studies is known as MSRVTT-1K, which consists of 9,000 videos for training and 1,000 videos for validation. Our main experiments and ablation studies are conducted on this split.

\textbf{DiDeMo~\cite{hendricks18emnlp}} dataset includes 10,000 videos. It allocates 8,395, 1,065 and 1,004 videos to training, validation and test set correspondingly.

\textbf{ActivityNet~\cite{caba2015activitynet}} contains 20,000 videos. Following the settings in \cite{eercf, clipvip,Zhang2018CrossModalAH}, we report the results on the validation set 1, which consists of 4,900 videos.

\textbf{VATEX~\cite{Wang2019VaTeXAL}} is another critical dataset in our study including 34,991 videos. Adhering to the division methods in \cite{Chen2020FineGrainedVR, ts2net}, we use 25,991 videos for training, 1,500 for validation, and another 1,500 for testing.

\textbf{MSVD~\cite{msvd}} dataset encompasses 1,970 video clips. For experimental consistency, we divided the dataset into 1,200 videos for training, 100 for validation, and 670 for testing.

\textbf{LSMDC~ \cite{lsmdc}} dataset is sourced from movie clips,  consisting of 101,079, 7,408 and 1,000 videos for training, validation and test set correspondingly.

Our experimental results are positioned to demonstrate the robustness and effectiveness of TokenBinder across diverse datasets and retrieval challenges.

\subsection{Evaluation Metric} 

In our evaluation, we employ the most widely used metrics in video text retrieval to assess the performance of our method on the aforementioned datasets. These metrics include Recall at 1 (R@1), Recall at 5 (R@5), and Recall at 10 (R@10). Higher values in these metrics indicate better performance of the retrieval system, as they reflect the percentage of correct items found in the top 1, 5, and 10 positions, respectively. Additionally, we report on the Median Recall (MdR) and the Mean Recall (MeanR). Median Recall provides the median rank position of the first relevant document in the result list, which serves as a robust indicator of the central tendency of retrieval effectiveness, minimizing the impact of outliers. Mean Recall calculates the average rank position of all relevant documents across the test queries, offering a comprehensive measure of the overall retrieval performance.

\subsection{Training Details}

Our model is built upon the CLIP-VIP \cite{clipvip} framework, and we utilize Pytorch version 1.8.0. Consistent with the practices observed in \cite{clip4clip} \cite{ts2net}, we begin by loading pretrained weights from CLIP to initialize our vision and text encoders. The query indicators are initialized using Kaiming normalization. For optimization, we employ the AdamW optimizer with a total batch size of 128. The learning rate is set to 1e-4 for the Focused-view fusion module and 1e-6 for other components, with a weight decay factor of 0.2. Given the distinct characteristics of each dataset, we set the training epochs for MSR-VTT, DiDeMo, ActivityNet, MSVD, VATEX and LSMDC at 5, 20, 20, 20, 20 and 10, respectively. Additionally, we conducted an ablation study on the MSR-VTT-1K-Test split to further assess the impact of various components and settings on the performance of our retrieval system.

\begin{table}[tbp]

\caption{Ablation study on different components}

\label{tab:ablation_components}
\centering
\small
\setlength{\tabcolsep}{8pt}
\begin{tabular}{l|l|cccc}
\toprule[1pt]
\multirow{3}{*}{\textbf{\rotatebox{90}{Compo.}}}   & \textbf{\begin{tabular}[c]{@{}l@{}}+ query\\ indicators\end{tabular}} &              & \ding{52}            & \ding{52}          & \ding{52}          \\
                                   & \textbf{\begin{tabular}[c]{@{}l@{}}+ stage1\\ scores\end{tabular}}    &              &               & \ding{52}          & \ding{52}           \\
                                   & \textbf{\begin{tabular}[c]{@{}l@{}}+ gumbel \\ softmax\end{tabular}}  &              &               &              & \ding{52}       \\ \hline
\multirow{10}{*}{\textbf{\rotatebox{90}{Metrics}}} & \textit{t2v} R@1                                                      & 50.0         & 51.8          & 52.0         & \textbf{52.5} \\
                                   & \textit{t2v} R@5                                                     & 74.3         & \textbf{75.8} & 75.7         & 75.3          \\
                                   & \textit{t2v} R@10                                                     & 83.5         & 83.6          & 83.8         & \textbf{84.2} \\
                                   & \textit{t2v} MdR                                                      & \textbf{1.0} & \textbf{1.0}  & \textbf{1.0} & \textbf{1.0}  \\
                                   & \textit{t2v} MnR                                                      & 13.2         & 13.6          & 13.5         & \textbf{12.9} \\ \cline{2-6} 
                                   & \textit{v2t} R@1                                                      & 48.0         & 51.3          & 51.2         & \textbf{52.0} \\
                                   & \textit{v2t} R@5                                                      & 74.8         & 74.9          & 75.4         & \textbf{75.5} \\
                                   & \textit{v2t} R@10                                                     & 83.2         & 83.9          & 84.3         & \textbf{84.6} \\
                                   & \textit{v2t} MdR                                                      & 2.0          & 2.0           & \textbf{1.0} & \textbf{1.0}  \\
                                   & \textit{v2t} MnR                                                      & 9.3          & 9.2           & 9.1          & \textbf{9.1}  \\ \bottomrule[1pt]
\end{tabular}
\end{table}

\begin{table}[tbp]
\caption{Abalation study on different number of query indicators}

\label{tab:ablation_number_query_indicator}
\centering
\resizebox{0.48\textwidth}{!}
{
\begin{tabular}{c|ccc|ccc}
\toprule[1.2pt]
\multirow{2}{*}{{\begin{tabular}[c]{@{}c@{}}Query \\ Indicators\end{tabular}}} & \multicolumn{3}{c|}{\textit{t2v}}   & \multicolumn{3}{c}{\textit{v2t}}  \\
                                                                                  & R@1  $\uparrow$         & R@5  $\uparrow$         & MnR     $\downarrow$      & R@1      $\uparrow$     & R@5    $\uparrow$  & MnR  $\downarrow$      \\ \hline
2                                                                                 & 50.6          & \textbf{75.4} & 13.2          & 50.3          & 75.3          & \textbf{8.9} \\
3                                                                                 & 51.2          & \textbf{75.4} & 13.9          & 51.1          & 75.4          & 9.3          \\
4                                                                                 & \textbf{52.5} & 75.3          & \textbf{12.9} & \textbf{52.0} & \textbf{75.5} & 9.1          \\
5                                                                                 & 52.0          & 74.6          & 13.8          & 50.4          & 73.6          & 8.8          \\
6                                                                                 & 50.2          & 73.8          & 13.9          & 49.6          & 74.5          & 9.5          \\ \bottomrule[1.2pt]
\end{tabular}
}
\end{table}

\subsection{Comparison Results}
In our experimental evaluation on the MSRVTT-1K-Test dataset, as summarized in Table ~\ref{tab:msrvtt1k}, we segmented the analysis into three distinct sections. In the non-CLIP segment, our model demonstrates competitive performance. For the CLIP-based models, we specifically highlight the results obtained with the ViT-B/32 and ViT-B/16 as backbone architectures. Notably, the results for the CLIP-ViP \cite{clipvip} model were reproduced by its open-source codebase. Our approach, underpinned by a ViT-B/32 backbone, surpasses the existing methods across the majority of the evaluated metrics. The implementation with a ViT-B/16 backbone further amplifies this superiority. This superiority is attributed to our method's robust feature extraction and cross-modality aggregation mechanism, which collectively enhance the alignment of textual and visual representations for more accurate retrieval performance.

Our extensive experimental evaluation extends beyond the predominant MSRVTT-1K dataset, encompassing a variety of datasets, as detailed in Table \ref{tab:multi_dataset}. The results on ActivityNet, DiDeMo, VATEX, MSVD alongside the LSMDC dataset, underscore the versatility and consistent performance of our TokenBinder model.

\begin{table}[tbp]
\caption{Abalation study on top-$k$ selection}

\label{tab:ablation_topk_selection}
\centering
\small
\resizebox{0.48\textwidth}{!}
{
\begin{tabular}{c|cccc|cccc}
\toprule[1.2pt]
\multirow{2}{*}{\textbf{\begin{tabular}[c]{@{}c@{}}Top\\ @$k$\end{tabular}}} & \multicolumn{4}{c|}{\textit{t2v} Retrieval}                  & \multicolumn{4}{c}{\textit{v2t} Retrieval}                  \\
                                                                           & R@1           & R@5           & R@10          & MnR           & R@1           & R@5           & R@10          & MnR          \\ \hline
5                                                                          & 52.1          & 75.1          & 84.1          & 13.0          & 52.0          & 75.4          & 84.3          & \textbf{9.1} \\
10                                                                         & \textbf{52.5} & 75.3          & 84.2          & \textbf{12.9} & \textbf{52.0} & 75.5          & \textbf{84.6} & \textbf{9.1} \\
15                                                                         & 52.4          & 75.5          & 84.2          & \textbf{12.9} & 51.7          & \textbf{75.7} & \textbf{84.6} & \textbf{9.1} \\
20                                                                         & 52.4          & 75.5          & 84.3          & 13.0          & 52.1          & 75.6          & 84.6          & \textbf{9.1} \\
40                                                                         & 51.3          & \textbf{76.5} & \textbf{84.5} & 14.3          & 51.7          & 75.6          & 84.4          & 10.0         \\ \bottomrule[1.2pt]
\end{tabular}
}

\end{table}

\begin{table}[tbp]

\caption{Abalation study on different number of focused-view inter-modality cross attention blocks}

\label{tab:ablation_cma_blocks}
\centering
\small
\resizebox{0.4\textwidth}{!}
{
\begin{tabular}{c|ccc|ccc}
\toprule[1.2pt]
\multirow{2}{*}{{\begin{tabular}[c]{@{}c@{}}Num of\\ Blocks\end{tabular}}} & \multicolumn{3}{c|}{{\textit{t2v} Retrieval}}   & \multicolumn{3}{c}{{\textit{v2t} Retrieval}}   \\
                                                                                   & R@1           & R@5           & MnR           & R@1           & R@5           & MnR          \\ \hline
1                                                                                  & \textbf{52.5} & 75.3          & \textbf{12.9} & \textbf{52.0} & 75.5          & \textbf{9.1} \\
2                                                                                  & \textbf{52.5} & \textbf{75.4} & 13.2          & \textbf{52.0} & 75.8          & 9.2          \\
3                                                                                  & 52.3          & \textbf{75.4} & 13.8          & 51.1          & \textbf{77.1} & 9.2          \\ \bottomrule[1.2pt]
\end{tabular}
}

\end{table}

\begin{figure*}[ht]
    \centering
    \includegraphics[width=\linewidth]{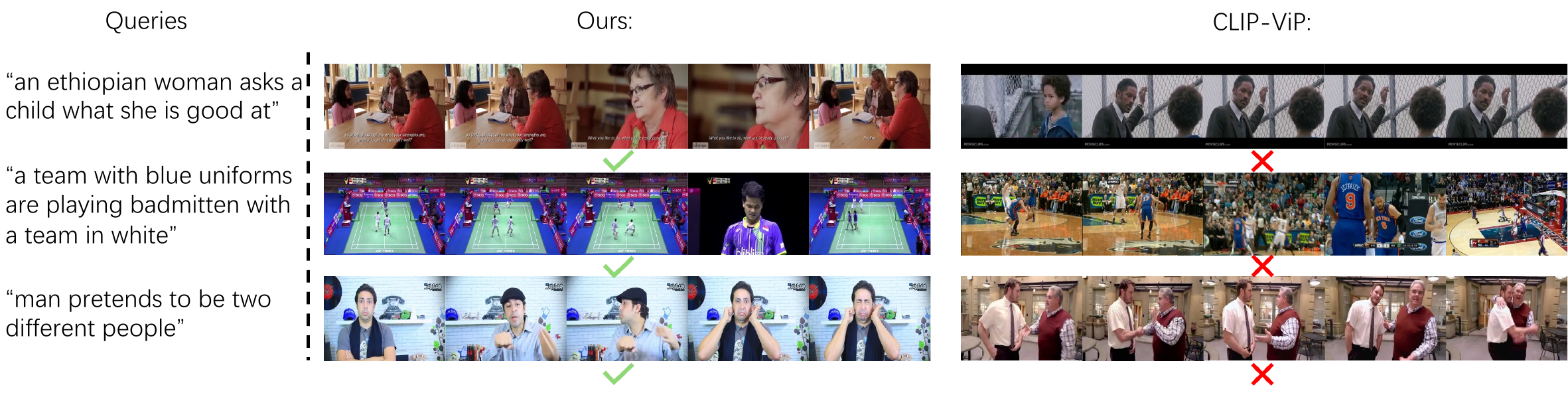}

    \caption{\textbf{Example of Text-to-Video Retrieval produced by TokenBinder and CLIP-ViP. }
    A green tick signifies successful retrieval of the accurate video, while a red cross denotes an erroneous retrieval outcome.}

    \label{fig:case_study}
\end{figure*}

\subsection{Ablation Study}

\noindent \textbf{Impact of different components.} As presented in Table~\ref{tab:ablation_components}, we study the impacts of various components, \ie query indicators, add stage-1 scores, and the Gumbel softmax, on performance. 
Adding query indicators yields marginal improvements, while stage-1 scores further enhance results. Gumbel softmax significantly boosts R@10 for both \textit{t2v} and \textit{v2t}, indicating its role in fine-grained feature discrimination. The combined integration of all components achieves the highest Recall@1 for both retrieval directions, showing their synergistic effect on retrieval precision.

\noindent \textbf{Impact of number of query indicators.} As illustrated in Table~\ref{tab:ablation_number_query_indicator}, we evaluate the influence of the number of query indicators on the efficacy of our method. Note that a minimum of two query indicators is necessary for focused-view retrieval. Increasing the number of indicators from two to six improves R@1 in \textit{t2v}, with the best performance at four tokens, followed by a slight decrease. 
Meanwhile, in the \textit{v2t} retrieval context, an uptick in R@1 is perceptible with two query indicators, followed by a plateau, and then a slight diminution with six tokens, implying a nuanced upper bound for token quantity. 

\noindent \textbf{Impact of top-$k$ selections.} As shown in Table~\ref{tab:ablation_topk_selection}, we investigate the impact of varying the top-$k$ selection values. For both the retrieval of \textit{t2v} and \textit{v2t}, the results demonstrate that R@1 marginally peaks at k=10, indicating a slight preference for intermediate set sizes of candidates. 
Notably, as k increases to 40, R@5 and R@10 metrics exhibit a subtle improvement in \textit{t2v} retrieval, suggesting that a more comprehensive review of the candidate could be beneficial.

\noindent \textbf{Impact of number of Focused-view attention blocks.} In Table \ref{tab:ablation_cma_blocks}, we examine the effect of varying the number of inter-modality cross attention blocks of our model. Our study considers configurations with one, two, and three blocks. The results indicate a marginal increase in both R@5 and MnR as the number of blocks increases from one to three for \textit{t2v} retrieval. However, due to the minimal gains and the computational cost of additional blocks, we select a single cross-attention block in the final model to balance performance and efficiency.

\subsection{Case Study}

As demonstrated in Figure~\ref{fig:case_study}, upon examining the provided case study, we analyze contrasting instances to delineate the efficacy of the TokenBinder in discerning complex video-text correlations. For instance, our model accurately matches the query "an Ethiopian woman asks a child what she is good at" with the corresponding video, showcasing the model's nuanced understanding of the scene, as opposed to CLIP-ViP's erroneous retrieval. In another case, TokenBinder effectively identifies the correct video for "a team with blue uniforms are playing badminton with a team in white", which demonstrates the model's capability to recognize and differentiate between subtle visual details and activities. Furthermore, our framework successfully disambiguates the challenging query "man pretends to be two different people", illustrating its adeptness at interpreting complex human behaviors and interactions within videos. These case studies exemplify the advanced interpretative performance of TokenBinder. 

\section{Conclusion}
In this paper, we introduced TokenBinder, utilizing a novel one-to-many alignment paradigm to enhance text-video retrieval. Extensive experiments on six  datasets confirmed that TokenBinder outperforms existing state-of-the-art methods, demonstrating robust fine-grained alignment and effective bridging of intra- and inter-modality information gaps. These findings underscore TokenBinder's potential to advance the field of text-video retrieval, offering a more accurate and nuanced retrieval framework.

\section*{Acknowledgements}
 This work is supported by Australian Research Council Discovery Project DP230101753, and CSIRO’s Science Leader Project R-91559.   

\clearpage
{\small
\bibliographystyle{ieee_fullname}
\bibliography{egbib}

\begin{thebibliography}{10}\itemsep=-1pt

\bibitem{frozen}
Max Bain, Arsha Nagrani, G{\"{u}}l Varol, and Andrew Zisserman.
\newblock Frozen in time: {A} joint video and image encoder for end-to-end retrieval.
\newblock In {\em 2021 {IEEE/CVF} International Conference on Computer Vision, {ICCV} 2021, Montreal, QC, Canada, October 10-17, 2021}, pages 1708--1718. {IEEE}, 2021.

\bibitem{Bogolin2021CrossMR}
Simion-Vlad Bogolin, Ioana Croitoru, Hailin Jin, Yang Liu, and Samuel Albanie.
\newblock Cross modal retrieval with querybank normalisation.
\newblock {\em 2022 IEEE/CVF Conference on Computer Vision and Pattern Recognition (CVPR)}, pages 5184--5195, 2021.

\bibitem{msvd}
David~L. Chen and William~B. Dolan.
\newblock Collecting highly parallel data for paraphrase evaluation.
\newblock In {\em Proceedings of the 49th Annual Meeting of the Association for Computational Linguistics (ACL-2011)}, Portland, OR, June 2011.

\bibitem{Chen2020FineGrainedVR}
Shizhe Chen, Yida Zhao, Qin Jin, and Qi Wu.
\newblock Fine-grained video-text retrieval with hierarchical graph reasoning.
\newblock {\em 2020 IEEE/CVF Conference on Computer Vision and Pattern Recognition (CVPR)}, pages 10635--10644, 2020.

\bibitem{Chen2023TaggingBA}
Yizhen Chen, Jie Wang, Lijian Lin, Zhongang Qi, Jin Ma, and Ying Shan.
\newblock Tagging before alignment: Integrating multi-modal tags for video-text retrieval.
\newblock {\em ArXiv}, abs/2301.12644, 2023.

\bibitem{camoe}
Xing Cheng, Hezheng Lin, Xiangyu Wu, Fan Yang, and Dong Shen.
\newblock Improving video-text retrieval by multi-stream corpus alignment and dual softmax loss.
\newblock {\em CoRR}, abs/2109.04290, 2021.

\bibitem{teachtext}
Ioana Croitoru, Simion-Vlad Bogolin, Yang Liu, Samuel Albanie, Marius Leordeanu, Hailin Jin, and Andrew Zisserman.
\newblock Teachtext: Crossmodal generalized distillation for text-video retrieval.
\newblock {\em 2021 IEEE/CVF International Conference on Computer Vision (ICCV)}, pages 11563--11573, 2021.

\bibitem{prompt_switch}
Chaorui Deng, Qi Chen, Pengda Qin, Da Chen, and Qi Wu.
\newblock Prompt switch: Efficient {CLIP} adaptation for text-video retrieval.
\newblock In {\em {IEEE/CVF} International Conference on Computer Vision, {ICCV} 2023, Paris, France, October 1-6, 2023}, pages 15602--15612. {IEEE}, 2023.

\bibitem{vit}
Alexey Dosovitskiy, Lucas Beyer, Alexander Kolesnikov, Dirk Weissenborn, Xiaohua Zhai, Thomas Unterthiner, Mostafa Dehghani, Matthias Minderer, Georg Heigold, Sylvain Gelly, Jakob Uszkoreit, and Neil Houlsby.
\newblock An image is worth 16x16 words: Transformers for image recognition at scale.
\newblock {\em ArXiv}, abs/2010.11929, 2020.

\bibitem{caba2015activitynet}
Bernard~Ghanem Fabian Caba~Heilbron, Victor~Escorcia and Juan~Carlos Niebles.
\newblock Activitynet: A large-scale video benchmark for human activity understanding.
\newblock In {\em Proceedings of the IEEE Conference on Computer Vision and Pattern Recognition}, pages 961--970, 2015.

\bibitem{violet}
Tsu-Jui Fu, Linjie Li, Zhe Gan, Kevin Lin, William~Yang Wang, Lijuan Wang, and Zicheng Liu.
\newblock Violet : End-to-end video-language transformers with masked visual-token modeling.
\newblock {\em ArXiv}, abs/2111.12681, 2021.

\bibitem{MMT}
Valentin Gabeur, Chen Sun, Karteek Alahari, and Cordelia Schmid.
\newblock Multi-modal transformer for video retrieval.
\newblock In Andrea Vedaldi, Horst Bischof, Thomas Brox, and Jan{-}Michael Frahm, editors, {\em Computer Vision - {ECCV} 2020 - 16th European Conference, Glasgow, UK, August 23-28, 2020, Proceedings, Part {IV}}, volume 12349 of {\em Lecture Notes in Computer Science}, pages 214--229. Springer, 2020.

\bibitem{clip2tv}
Zijian Gao, Jingyu Liu, Sheng Chen, Dedan Chang, Hao Zhang, and Jinwei Yuan.
\newblock {CLIP2TV:} an empirical study on transformer-based methods for video-text retrieval.
\newblock {\em CoRR}, abs/2111.05610, 2021.

\bibitem{xpool}
Satya~Krishna Gorti, No{\"{e}}l Vouitsis, Junwei Ma, Keyvan Golestan, Maksims Volkovs, Animesh Garg, and Guangwei Yu.
\newblock X-pool: Cross-modal language-video attention for text-video retrieval.
\newblock In {\em {IEEE/CVF} Conference on Computer Vision and Pattern Recognition, {CVPR} 2022, New Orleans, LA, USA, June 18-24, 2022}, pages 4996--5005. {IEEE}, 2022.

\bibitem{hendricks18emnlp}
Lisa~Anne Hendricks, Oliver Wang, Eli Shechtman, Josef Sivic, Trevor Darrell, and Bryan Russell.
\newblock Localizing moments in video with temporal language.
\newblock In {\em Empirical Methods in Natural Language Processing (EMNLP)}, 2018.

\bibitem{Jang2016CategoricalRW}
Eric Jang, Shixiang~Shane Gu, and Ben Poole.
\newblock Categorical reparameterization with gumbel-softmax.
\newblock {\em ArXiv}, abs/1611.01144, 2016.

\bibitem{Kaufman2016TemporalTA}
Dotan Kaufman, Gil Levi, Tal Hassner, and Lior Wolf.
\newblock Temporal tessellation: A unified approach for video analysis.
\newblock {\em 2017 IEEE International Conference on Computer Vision (ICCV)}, pages 94--104, 2016.

\bibitem{Kiros2014UnifyingVE}
Ryan Kiros, Ruslan Salakhutdinov, and Richard~S. Zemel.
\newblock Unifying visual-semantic embeddings with multimodal neural language models.
\newblock {\em ArXiv}, abs/1411.2539, 2014.

\bibitem{prost}
Pandeng Li, Chen{-}Wei Xie, Liming Zhao, Hongtao Xie, Jiannan Ge, Yun Zheng, Deli Zhao, and Yongdong Zhang.
\newblock Progressive spatio-temporal prototype matching for text-video retrieval.
\newblock In {\em {IEEE/CVF} International Conference on Computer Vision, {ICCV} 2023, Paris, France, October 1-6, 2023}, pages 4077--4087. {IEEE}, 2023.

\bibitem{hit}
Song Liu, Haoqi Fan, Shengsheng Qian, Yiru Chen, Wenkui Ding, and Zhongyuan Wang.
\newblock Hit: Hierarchical transformer with momentum contrast for video-text retrieval.
\newblock In {\em 2021 {IEEE/CVF} International Conference on Computer Vision, {ICCV} 2021, Montreal, QC, Canada, October 10-17, 2021}, pages 11895--11905. {IEEE}, 2021.

\bibitem{CE}
Yang Liu, Samuel Albanie, Arsha Nagrani, and Andrew Zisserman.
\newblock Use what you have: Video retrieval using representations from collaborative experts.
\newblock In {\em 30th British Machine Vision Conference 2019, {BMVC} 2019, Cardiff, UK, September 9-12, 2019}, page 279. {BMVA} Press, 2019.

\bibitem{ts2net}
Yuqi Liu, Pengfei Xiong, Luhui Xu, Shengming Cao, and Qin Jin.
\newblock Ts2-net: Token shift and selection transformer for text-video retrieval.
\newblock In Shai Avidan, Gabriel~J. Brostow, Moustapha Ciss{\'{e}}, Giovanni~Maria Farinella, and Tal Hassner, editors, {\em Computer Vision - {ECCV} 2022 - 17th European Conference, Tel Aviv, Israel, October 23-27, 2022, Proceedings, Part {XIV}}, volume 13674 of {\em Lecture Notes in Computer Science}, pages 319--335. Springer, 2022.

\bibitem{Lu2023UniAdapterUP}
Haoyu Lu, Mingyu Ding, Yuqi Huo, Guoxing Yang, Zhiwu Lu, Masayoshi Tomizuka, and Wei Zhan.
\newblock Uniadapter: Unified parameter-efficient transfer learning for cross-modal modeling.
\newblock {\em ArXiv}, abs/2302.06605, 2023.

\bibitem{clip4clip}
Huaishao Luo, Lei Ji, Ming Zhong, Yang Chen, Wen Lei, Nan Duan, and Tianrui Li.
\newblock Clip4clip: An empirical study of {CLIP} for end to end video clip retrieval and captioning.
\newblock {\em Neurocomputing}, 508:293--304, 2022.

\bibitem{xclip}
Yiwei Ma, Guohai Xu, Xiaoshuai Sun, Ming Yan, Ji Zhang, and Rongrong Ji.
\newblock {X-CLIP:} end-to-end multi-grained contrastive learning for video-text retrieval.
\newblock In Jo{\~{a}}o Magalh{\~{a}}es, Alberto~Del Bimbo, Shin'ichi Satoh, Nicu Sebe, Xavier Alameda{-}Pineda, Qin Jin, Vincent Oria, and Laura Toni, editors, {\em {MM} '22: The 30th {ACM} International Conference on Multimedia, Lisboa, Portugal, October 10 - 14, 2022}, pages 638--647. {ACM}, 2022.

\bibitem{Maddison2016TheCD}
Chris~J. Maddison, Andriy Mnih, and Yee~Whye Teh.
\newblock The concrete distribution: A continuous relaxation of discrete random variables.
\newblock {\em ArXiv}, abs/1611.00712, 2016.

\bibitem{Park2022NormalizedCL}
Yookoon Park, Mahmoud Azab, Bo Xiong, Seungwhan Moon, Florian Metze, Gourab Kundu, and Kirmani Ahmed.
\newblock Normalized contrastive learning for text-video retrieval.
\newblock {\em ArXiv}, abs/2212.11790, 2022.

\bibitem{support_set}
Mandela Patrick, Po{-}Yao Huang, Yuki~Markus Asano, Florian Metze, Alexander~G. Hauptmann, Jo{\~{a}}o~F. Henriques, and Andrea Vedaldi.
\newblock Support-set bottlenecks for video-text representation learning.
\newblock In {\em 9th International Conference on Learning Representations, {ICLR} 2021, Virtual Event, Austria, May 3-7, 2021}. OpenReview.net, 2021.

\bibitem{pollitt2012method}
Alastair Pollitt.
\newblock The method of adaptive comparative judgement.
\newblock {\em Assessment in Education: principles, policy \& practice}, 19(3):281--300, 2012.

\bibitem{clip_straight}
Jes{\'{u}}s~Andr{\'{e}}s Portillo{-}Quintero, Jos{\'{e}}~Carlos Ortiz{-}Bayliss, and Hugo Terashima{-}Mar{\'{\i}}n.
\newblock A straightforward framework for video retrieval using {CLIP}.
\newblock In Edgar Roman{-}Rangel, {\'{A}}ngel Fernando~Kuri Morales, Jos{\'{e}} Francisco~Mart{\'{\i}}nez Trinidad, Jes{\'{u}}s~Ariel Carrasco{-}Ochoa, and Jos{\'{e}}~Arturo Olvera{-}L{\'{o}}pez, editors, {\em Pattern Recognition - 13th Mexican Conference, {MCPR} 2021, Mexico City, Mexico, June 23-26, 2021, Proceedings}, volume 12725 of {\em Lecture Notes in Computer Science}, pages 3--12. Springer, 2021.

\bibitem{clip}
Alec Radford, Jong~Wook Kim, Chris Hallacy, Aditya Ramesh, Gabriel Goh, Sandhini Agarwal, Girish Sastry, Amanda Askell, Pamela Mishkin, Jack Clark, et~al.
\newblock Learning transferable visual models from natural language supervision.
\newblock In {\em International conference on machine learning}, pages 8748--8763. PMLR, 2021.

\bibitem{rao2022denseclip}
Yongming Rao, Wenliang Zhao, Guangyi Chen, Yansong Tang, Zheng Zhu, Guan Huang, Jie Zhou, and Jiwen Lu.
\newblock Denseclip: Language-guided dense prediction with context-aware prompting.
\newblock In {\em Proceedings of the IEEE/CVF conference on computer vision and pattern recognition}, pages 18082--18091, 2022.

\bibitem{lsmdc}
Anna Rohrbach, Atousa Torabi, Marcus Rohrbach, Niket Tandon, Chris Pal, Hugo Larochelle, Aaron Courville, and Bernt Schiele.
\newblock Movie description.
\newblock {\em International Journal of Computer Vision}, 2017.

\bibitem{Shu2022MaskedCP}
Fangxun Shu, Biaolong Chen, Yue Liao, Ke Gao, Shuwen Xiao, Wenyu Sun, Xiaobo Li, Yousong Zhu, Jinqiao Wang, and Si Liu.
\newblock Masked contrastive pre-training for efficient video-text retrieval.
\newblock {\em ArXiv}, abs/2212.00986, 2022.

\bibitem{sweller1988cognitive}
John Sweller.
\newblock Cognitive load during problem solving: Effects on learning.
\newblock {\em Cognitive science}, 12(2):257--285, 1988.

\bibitem{eercf}
Kaibin Tian, Yanhua Cheng, Yi Liu, Xinglin Hou, Quan Chen, and Han Li.
\newblock Towards efficient and effective text-to-video retrieval with coarse-to-fine visual representation learning.
\newblock In {\em AAAI Conference on Artificial Intelligence}, 2024.

\bibitem{Torabi2016LearningLE}
Atousa Torabi, Niket Tandon, and Leonid Sigal.
\newblock Learning language-visual embedding for movie understanding with natural-language.
\newblock {\em ArXiv}, abs/1609.08124, 2016.

\bibitem{vaswani2017attention}
Ashish Vaswani, Noam Shazeer, Niki Parmar, Jakob Uszkoreit, Llion Jones, Aidan~N Gomez, {\L}ukasz Kaiser, and Illia Polosukhin.
\newblock Attention is all you need.
\newblock {\em Advances in neural information processing systems}, 30, 2017.

\bibitem{Wang2021ObjectawareVP}
Alex Wang, Yixiao Ge, Guanyu Cai, Rui Yan, Xudong Lin, Ying Shan, Xiaohu Qie, and Mike~Zheng Shou.
\newblock Object-aware video-language pre-training for retrieval.
\newblock {\em 2022 IEEE/CVF Conference on Computer Vision and Pattern Recognition (CVPR)}, pages 3303--3312, 2021.

\bibitem{drl}
Qiang Wang, Yanhao Zhang, Yun Zheng, Pan Pan, and Xian{-}Sheng Hua.
\newblock Disentangled representation learning for text-video retrieval.
\newblock {\em CoRR}, abs/2203.07111, 2022.

\bibitem{Wang2019VaTeXAL}
Xin~Eric Wang, Jiawei Wu, Junkun Chen, Lei Li, Yuan fang Wang, and William~Yang Wang.
\newblock Vatex: A large-scale, high-quality multilingual dataset for video-and-language research.
\newblock {\em 2019 IEEE/CVF International Conference on Computer Vision (ICCV)}, pages 4580--4590, 2019.

\bibitem{ucofia}
Ziyang Wang, Yi{-}Lin Sung, Feng Cheng, Gedas Bertasius, and Mohit Bansal.
\newblock Unified coarse-to-fine alignment for video-text retrieval.
\newblock In {\em {IEEE/CVF} International Conference on Computer Vision, {ICCV} 2023, Paris, France, October 1-6, 2023}, pages 2804--2815. {IEEE}, 2023.

\bibitem{xu2016msr-vtt}
Jun Xu, Tao Mei, Ting Yao, and Yong Rui.
\newblock Msr-vtt: A large video description dataset for bridging video and language.
\newblock IEEE International Conference on Computer Vision and Pattern Recognition (CVPR), June 2016.

\bibitem{clipvip}
Hongwei Xue, Yuchong Sun, Bei Liu, Jianlong Fu, Ruihua Song, Houqiang Li, and Jiebo Luo.
\newblock Clip-vip: Adapting pre-trained image-text model to video-language representation alignment.
\newblock {\em CoRR}, abs/2209.06430, 2022.

\bibitem{Yang2021TACoTC}
Jianwei Yang, Yonatan Bisk, and Jianfeng Gao.
\newblock Taco: Token-aware cascade contrastive learning for video-text alignment.
\newblock {\em 2021 IEEE/CVF International Conference on Computer Vision (ICCV)}, pages 11542--11552, 2021.

\bibitem{dgl}
Xiangpeng Yang, Linchao Zhu, Xiaohan Wang, and Yi Yang.
\newblock {DGL:} dynamic global-local prompt tuning for text-video retrieval.
\newblock In Michael~J. Wooldridge, Jennifer~G. Dy, and Sriraam Natarajan, editors, {\em Thirty-Eighth {AAAI} Conference on Artificial Intelligence, {AAAI} 2024, Thirty-Sixth Conference on Innovative Applications of Artificial Intelligence, {IAAI} 2024, Fourteenth Symposium on Educational Advances in Artificial Intelligence, {EAAI} 2014, February 20-27, 2024, Vancouver, Canada}, pages 6540--6548. {AAAI} Press, 2024.

\bibitem{Yu2018AJS}
Youngjae Yu, Jongseok Kim, and Gunhee Kim.
\newblock A joint sequence fusion model for video question answering and retrieval.
\newblock In {\em European Conference on Computer Vision}, 2018.

\bibitem{Zhang2018CrossModalAH}
Bowen Zhang, Hexiang Hu, and Fei Sha.
\newblock Cross-modal and hierarchical modeling of video and text.
\newblock In {\em European Conference on Computer Vision}, 2018.

\bibitem{Zhang2023MultieventVR}
Gengyuan Zhang, Jisen Ren, Jindong Gu, and Volker Tresp.
\newblock Multi-event video-text retrieval.
\newblock {\em 2023 IEEE/CVF International Conference on Computer Vision (ICCV)}, pages 22056--22066, 2023.

\bibitem{centerclip}
Shuai Zhao, Linchao Zhu, Xiaohan Wang, and Yi Yang.
\newblock Centerclip: Token clustering for efficient text-video retrieval.
\newblock In Enrique Amig{\'{o}}, Pablo Castells, Julio Gonzalo, Ben Carterette, J.~Shane Culpepper, and Gabriella Kazai, editors, {\em {SIGIR} '22: The 45th International {ACM} {SIGIR} Conference on Research and Development in Information Retrieval, Madrid, Spain, July 11 - 15, 2022}, pages 970--981. {ACM}, 2022.

\end{thebibliography}
}

\end{document}